\documentclass[journal]{IEEEtai}

\usepackage[colorlinks,urlcolor=blue,linkcolor=blue,citecolor=blue]{hyperref}

\usepackage{color,array}

\usepackage{graphicx}
\usepackage{amsmath}
\usepackage{amssymb}
\usepackage{booktabs}
\usepackage{pifont}
\usepackage{tikz}
\newcommand{\tikzxmark}{%
\tikz[scale=0.23] {
    \draw[line width=0.7,line cap=round] (0,0) to [bend left=6] (1,1);
    \draw[line width=0.7,line cap=round] (0.2,0.95) to [bend right=3] (0.8,0.05);
}}
\newcommand{\tikzcmark}{%
\tikz[scale=0.23] {
    \draw[line width=0.7,line cap=round] (0.25,0) to [bend left=10] (1,1);
    \draw[line width=0.8,line cap=round] (0,0.35) to [bend right=1] (0.23,0);
}}
\usepackage{color, colortbl}
\definecolor{Gray}{gray}{0.9}
\definecolor{LightCyan}{rgb}{0.88,1,1}
\definecolor{maroon}{cmyk}{0,0.87,0.68,0.32}\usepackage{multirow}
\newcolumntype{L}{>{\raggedright\arraybackslash}c}
\usepackage[linesnumbered,ruled]{algorithm2e}
\SetKwInput{KwInput}{Input}                % Set the Input
\SetKwInput{KwOutput}{Output}              % set the Output
\SetKwInOut{KwInit}{Initialization}
\begin{document}

\title{Curriculum Guided Domain Adaptation in the Dark}

\author{Chowdhury Sadman Jahan, and Andreas Savakis, \IEEEmembership{Senior Member, IEEE}
\thanks{Submitted for review on July 23, 2023. This research was partly supported by the Air Force Office of Scientific Research (AFOSR) under SBIR grant FA9550-22-P-0009 with Intelligent Fusion Technology and AFOSR grant FA9550-20-1-0039. }
\thanks{C. S. Jahan is with the Center for Imaging Science, Rochester Institute of Technology, Rochester, NY 14623 (e-mail: sj4654@rit.edu).}
\thanks{A. Savakis is with the Department of Computer Engineering, Kate Gleason College of Engineering, Rochester Institute of Technology, Rochester, NY 14623 (e-mail: andreas.savakis@rit.edu).}}

%\thanks{T. C. Author is with the Electrical Engineering Department, University of Colorado, Boulder, CO 80309 USA, on leave from the National Research Institute for Metals, Tsukuba, Japan (e-mail: author@nrim.go.jp).}
%%%%% Uncomment the three lines below before submission
%\thanks{This paragraph will include the Associate Editor who handled your paper.}}
%\markboth{Journal of IEEE Transactions on Artificial Intelligence, Vol. 00, No. 0, Month 2020}
%{Jahan \MakeLowercase{\textit{et al.}}: Curriculum Guided Domain Adaptation in the Dark}
%%%%% Uncomment the three lines above before submission

\maketitle

\begin{abstract}
Addressing the rising concerns of privacy and security, domain adaptation in the dark aims to adapt a black-box source trained model to an unlabeled target domain without access to any source data or source model parameters. 
The need for domain adaptation of black-box predictors becomes even more pronounced to protect intellectual property as deep learning based solutions are becoming increasingly commercialized.
Current methods distill noisy predictions on the target data obtained from the source model to the target model, and/or separate clean/noisy target samples before adapting using traditional noisy label learning algorithms. 
However, these methods do not utilize the easy-to-hard learning nature of the clean/noisy data splits.
Also, none of the existing methods are end-to-end, and require a separate fine-tuning stage and an initial warmup stage.
In this work, we present \textbf{C}urriculum \textbf{A}daptation for \textbf{B}lack-\textbf{B}ox (\textbf{CABB}) 
which provides a curriculum guided adaptation approach to gradually train the target model, first on  target data with high confidence (clean) labels, and later on target data with noisy labels. 
CABB utilizes Jensen-Shannon divergence as a better criterion for clean-noisy sample separation, compared to the traditional criterion  of cross entropy loss. 
Our method utilizes co-training of a dual-branch network to suppress error accumulation resulting from confirmation bias.
The proposed approach is end-to-end trainable and does not require any extra finetuning stage, unlike existing methods. 
Empirical results on standard domain adaptation datasets show that CABB outperforms existing state-of-the-art black-box DA models and is comparable to white-box domain adaptation models.
\end{abstract}

\begin{IEEEImpStatement}

In addition to preserving data privacy, commercialization of deep learning models has given rise to concerns about protecting proprietary rights. In order to alleviate these concerns, domain adaptation of black-box predictors (DABP) puts additional constraints on the already challenging domain adaptation problem by limiting access not only to the source data used for training, but also to the source model parameters during adaptation to the target domain. We take inspiration from noisy label learning and propose CABB as a curriculum guided domain adaptation approach for DABP using a dual-branch target model. Our clean-noisy sample separation process produces more accurate clean sample sets compared to the traditional sample filtering methods. The pseudolabels generated in CABB are also more robust. Unlike existing state-of-the-art, DABP methods, our model is end-to-end trainable, and outperforms other methods in all benchmarks we tested. Our method advances DABP and can have immediate impact to protect proprietary models and their training data during deployment and adaptation.

\end{IEEEImpStatement}

%\begin{IEEEImpStatement}
%The impact statement should not exceeed 150 words. This section offers an example that is expanded to have only and just 150 words to demonstrate the point. Here is an example on how to write an appropriate impact statement: Chatbots are a popular technology in online interaction. They reduce the load on human support teams and offer continuous 24-7 support to customers. However, recent usability research has demonstrated that 30\% of customers are unhappy with current chatbots due to their poor conversational capabilities and inability to emotionally engage customers. The natural language algorithms we introduce in this paper overcame these limitations. With a significant increase in user satisfaction to 92\% after adopting our algorithms, the technology is ready to support users in a wide variety of applications including government front shops, automatic tellers, and the gaming industry. It could offer an alternative way of interaction for some physically disable users.
%\end{IEEEImpStatement}


\section{Introduction}
\label{sec:intro}

\begin{figure}
\begin{minipage}[b]{1\linewidth}
  \centering
    \includegraphics[width=.8\textwidth]{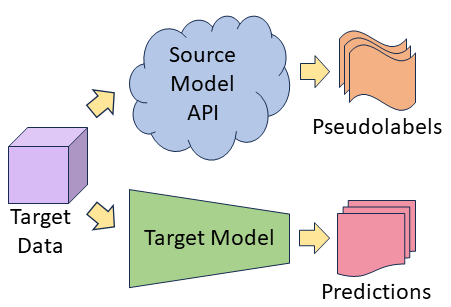}
\end{minipage}
    \caption{
    % Premise 
    Overview of domain adaptation for black-box predictors (DABP). The source model may only be accessed to generate pseudolabels for the target data, and these pseudolabels may be used to adapt another model on the target domain.}
    \label{fig:DABP}
\end{figure}

\IEEEPARstart{W}{ith} the availability of massive amounts of labelled image data, deep learning methods have made great progress in numerous computer vision tasks, such as classification, segmentation and object detection, among others. 
However, it is not feasible to collect, and annotate huge amounts of data for every new environment where a deep network model may be deployed. 
Unsupervised domain adaptation (UDA) is a special case of domain adaptation (DA) and transfer learning which aims to mitigate the domain gap that arises from deploying a model trained on labelled source data, to a new environment with unlabelled target data. 
Most of the existing UDA methods either adversarially align the labelled source data features, and unlabelled target data features \cite{ganin2016domain, hoffman2018cycada}, or minimize their distribution discrepancy \cite{torralba2011unbiased, tzeng2014deep, zellingercentral}. 
These methods require access to the source data during adaptation, and therefore cannot be applied when the source data is either unavailable, or cannot be shared due to privacy and security concerns. 
A newer, more efficient UDA paradigm, called Source-Free UDA \cite{shot, yang2021generalized} has recently emerged to address such cases, where the adaptation process, instead of the source data itself, utilizes only a model trained on the source data.
Such methods still fail to adequately alleviate data privacy and security concerns as 
% generative adversarial learning \cite{goodfellow2014generative} 
model attacks
may potentially retrieve the raw source data or corrupt the model. 
Moreover, with the commercialization of deep learning based solutions, companies may be reluctant to share their proprietary model parameters with the end users. 
These issues brought forth a newer UDA paradigm called domain adaptation of black-box predictors (DABP) that adapts without accessing neither the source data, nor the source model parameters \cite{liang2022dine}. 
Practically, a vendor can have the source trained model as an API in the cloud, and the end user can access the black-box source model to generate predictions for each unlabelled target instance to adapt on the target domain. 

Existing DABP methods transfer knowledge from the source trained model predictions to the target model, and then finetune the target model on the target data \cite{liang2022dine, yang2023divide}.
The approach in
\cite{yang2023divide} utilizes a noisy label learning (NLL) algorithm \cite{Li2020DivideMix} to separate the target domain into an easy-to-adapt subdomain with cleaner pseudolabels, and a hard-to-adapt subdomain with noisier pseudolables using low cross-entropy (CE) loss criterion as the separator \cite{han2018co}, and then applies supervised and semi-supervised learning strategies on the easy- and hard-to-adapt subdomains, respectively.

In this work, we propose \textit{\textbf{C}urriculum \textbf{A}daptation for \textbf{B}lack-\textbf{B}ox \textbf{(CABB)}} as an unsupervised domain adaptation framework for black-box predictors. 
We present Jensen-Shannon distance (JSD) as a better criterion to separate clean and noisy samples using pseudolabels generated by the source model. 
JSD can be modelled using a two-component Gaussian Mixture Model (GMM) where the distribution with the lower distance can be considered to be consisting of cleaner samples and that with the higher distance contains noisier samples. 
As opposed to traditional low loss criterion for clean-noisy separation, low JSD criterion produces a more conservative, but more accurate clean sample set. 
To reduce error accumulation from confirmation bias, CABB employs co-training \cite{han2018co, Li2020DivideMix} two identical networks and adapts one network on the clean-noisy separated sets generated by the other, and vice versa. 
CABB introduces a curriculum learning strategy to adaptively learn from the clean samples first, and the noisy samples later during the adaptation process. 
CABB foregoes the finetuning stage of existing methods by utilizing mutual information maximization \cite{shot,conda} within its curriculum, making it end-to-end adaptable. 
The main contributions of our work are as follows.

\begin{itemize}
\item We introduce CABB as a curriculum guided domain adaptation model that progressively learns from the clean target set and the noisy target set, while utilizing co-training of a dual-branch network to suppress error accumulation resulting from confirmation bias.
\item We identify Jensen-Shannon divergence loss as a better criterion than cross-entropy loss for separation of clean and noisy samples for DABP.
\item CABB incorporates mutual information maximization within its curriculum and makes the adaptation process end-to-end without the need for any separate finetuning stage.
\item CABB produces robust pseudolabels from the mean of an ensemble of predictions generated by the two branches of the network on a set of augmentations.
\end{itemize}

\section{Related Works}

\subsection{Unsupervised domain adaptation}
 
Domain gap or domain shift occurs when the data distribution of the training data (source domain) is considerably different from that of the testing data (target domain)  \cite{torralba2011unbiased}. 
Long \textit{et al.} \cite{long2015learning}, and Tzeng \textit{et al.} \cite{tzeng2014deep} proposed to mitigate this distribution shift by minimizing the maximum mean discrepancy (MMD) between the two distributions, while Zellinger \textit{et al.} \cite{zellingercentral} proposed to match the higher order central moments of source and target probability distributions, and thus minimize central moment discrepancy (CMD) for UDA. 
Sun and Saenko \cite{sun2016deep} devised Deep CORAL to minimize second-order distribution statistics to mitigate domain shift. 
Ganin \textit{et al.} \cite{ganin2016domain} utilized a domain discriminator module, and introduced gradient reversal layer (GRL) to adversarially align the two distributions. 
Many methods followed since then that have utilized adversarial alignment on the latent feature space. \cite{long2017deep, pei2018multi}. 
While \cite{ganin2016domain} uses a common encoder for the source and target data, Tzeng \textit{et al.} \cite{tzeng2017adversarial} proposed to decouple the encoders by first training an encoder and a classifier on the labelled source data, followed by training a separate target data encoder using a domain discriminator, and finally deploying the same source classifier as the target classifier.
Hoffman \textit{et al.} \cite{pmlr-v80-hoffman18a} produced source-like images using generative image-to-image translation \cite{zhu2017unpaired} and adversarially aligned source and target data distributions at the low-level or pixel-level.
Global domain-wise adversarial alignment however may cause loss of intrinsic target class discrimination in the embedding space, and lead to suboptimal performance. 
To preserve class-wise feature discrimination, Li \textit{et al.} \cite{li2019joint} simultaneously aligned the domain-wise and class-wise distributions across the source and target data by solving two complementary domain-specific and class-specific minimax problems. 
In a non-adversarial approach, Pan \textit{et al.} \cite{pan2019transferrable} proposed to calculate the source class prototypes for the labelled source data, and target class prototypes from the pseudo-labelled target data, and then enforce consistency on the prototypes in the embedding space.
Tang \textit{et al.} \cite{tang2020unsupervised} similarly bases structural domain similarity to enforce structural source regularization and conducts discriminative clustering of target data without any domain alignment.
Chen \textit{et al.} \cite{chen2020graph} introduced graph matching to formulate cross-domain adaptation, and minimized Wasserstein distance for entity-matching, and Gromov-Wasserstein distance for edge matching. 
In order to reduce negative transfer introduced by the target samples that are either near the source-data generated decision boundaries, or are far away from their corresponding class centers, Xu \textit{et al.} \cite{xu2020reliable} proposed a weighted optimal transport strategy to achieve a reliable precise-pair-wise optimal transport procedure for domain adaptation.

Although domain divergence minimization \cite{torralba2011unbiased, tzeng2014deep, zellingercentral}, adversarial adaptation \cite{ganin2016domain, hoffman2018cycada}, and optimal transport \cite{chen2020graph, xu2020reliable} are widely used techniques for UDA, they require access to both the source and target data during adaptation. 
Addressing situations where source data is unavailable, several source-free DA (SFDA) methods have been proposed recently. 
Chidlovskii \textit{et al.} \cite{chidlovskii2016domain} proposed to use a few source prototypes or representatives in place of the entire source data for semi-supervised domain adaptation.
Liang \textit{et al.} \cite{liang2019distant} proposed to conduct target adaptation using source-free distant supervision to iteratively find taret pseudo-labels, a domain invariant subspace where the source and target data centroids are only moderately shifted, and finally target centroids/prototypes by implementing an alternating minimization strategy. 
Liang \textit{et al.} \cite{shot} introduced SHOT as an SFDA framework which transfers the source hypothesis or classifier to the target model, and adapts via self-training with information maximization \cite{krause2010discriminative, shi2012information, hu2017learning} and class centroid-based pseudolabel refinement. 
Yang \textit{et al.} \cite{yang2021generalized} proposed G-SFDA which refines the pseudolabels further via consistency regularization among neighboring target samples. 
Ding \textit{et al.} \cite{ding2022source} introduced SFDA-DE which samples from an estimated source data distribution, and conducts contrastive alignment between the estimated source and target distributions.

\subsection{Black box domain adaptation}

\begin{figure*}
\begin{minipage}[b]{1.0\linewidth}
  \centering
    \includegraphics[width=0.8\textwidth]{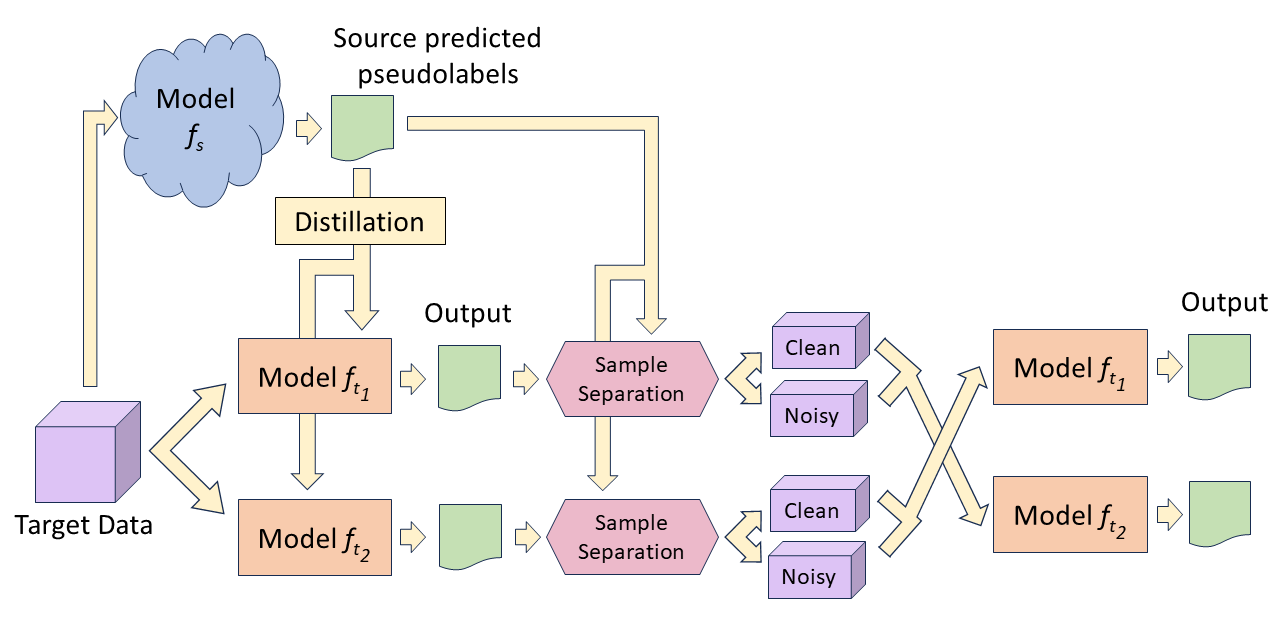}
\end{minipage}
    \caption{UDA pipeline in CABB. The target data is fed to the source model $f_s$ and the knowledge generated from $f_s$ is
    % distilled
    transferred to both target branches $f_{t_1}$ and $f_{t_2}$. The source predicted pseudolabels are also used to calculate JSD and produce clean-noisy sample sets. In subsequent co-training of $f_{t_1}$ and $f_{t_2}$, the samples sets created by one branch are used to update the other branch, using curriculum guided losses to progressively adapt to clean samples first, and the noisy samples later.}
    \label{fig:CABB model}
\end{figure*}

Extending the premise of SFDA further, Liang \textit{et al.} \cite{liang2022dine} introduced a newer paradigm of black box DA where, in addition to the source data, the source model parameters are also unavailable during adaptation. 
This new challenging scenario is important to protect intellectual property (source model parameters) from the end users. 
Liang \textit{et al.} proposed DINE which distills knowledge from the black-box source model to the target model in the first stage, followed by finetuning with target pseudolabels in the second stage. 
Yang \textit{et al.} \cite{yang2023divide} proposed BETA as a method that separates easy- and hard-to-learn pseudolabels using a conventional noisy label learning technique \cite{han2018co}, and applies a twin-network co-training strategy similar to \cite{Li2020DivideMix}, and adversarial alignment during adaptation. 

In this paper, we identify Jensen-Shannon distance (JSD) as a more appropriate criterion for clean-noisy sample separation for the unbounded noise rate in UDA, compared to traditional low CE loss for the bounded loss in NLL. We formulate a curriculum learning strategy to train the target model end-to-end with cleaner samples first, and progressively with noisy samples later. 

\section{Methodology}

%\begin{figure*}
%  \centering
%  \begin{subfigure}{0.65\linewidth}
    %\fbox{\rule{0pt}{2in} \rule{.9\linewidth}{0pt}}
%    \includegraphics[width=1\textwidth]{WACV 2024 image new final.png}
%    \caption{UDA pipeline in CABB.}
%    \label{fig:short-a}
%  \end{subfigure}
%  \hfill
%  \begin{subfigure}{0.32\linewidth}
%    %\fbox{\rule{0pt}{2in} \rule{.9\linewidth}{0pt}}
%    \includegraphics[width=1\textwidth]{WACV 2024 image 2 new final.png}
%    \caption{Ensemble-based pseudolabeling in CABB}
%    \label{fig:short-b}
%  \end{subfigure}
%  \caption{CABB for black-box domain adaptation.}
%  \label{fig:short}
%\end{figure*}

The black-box source model $f_s (\theta_s): \mathcal{X}_s \rightarrow \mathcal{Y}_s$ with model parameters $\theta_s$, maps the multiclass source data $x_s \in \mathcal{X}_s$ of source domain $\mathcal{D}_s$, to the label space $y_s \in \mathcal{Y}_s$. 
For DABP, we however do not have access to $\theta_s$, but only the hard predictions $(\hat{y_t} \in \mathcal{Y}_t) = f_s (\theta_s, x_t)$ from $f_s$ on the target data $x_t \in \mathcal{X}_t$ of target domain $\mathcal{D}_t$. 
There exists a domain shift between the source data distribution $\mathcal{D}_s$ and the target data distribution $\mathcal{D}_t$, while the label space is shared, i,e $\mathcal{Y}_s = \mathcal{Y}_t$. 
Due to this domain shift, a large number of predictions $\hat{y_t}$ may be incorrect and could result in a set of noisy pseudolabels generated by the source model.
%, producing a set of source model generated noisy pseudolabels.
Our objective for DA is to learn a mapping function $f_t (\theta_t): \mathcal{X}_t \rightarrow \mathcal{Y}_t$. 

Research has shown that when deep networks are trained with noisy labels, the resulting models tend to memorize the wrongly labelled samples owing to confirmation bias, as the training progresses \cite{zhang2017understanding}.
Furthermore, in regular training of a single-branch network with noisy labels, the error from one training mini-batch flows back into the network itself for the next mini-batch, and thus the error increasingly accumulates \cite{han2018co}.
In this work, during adaptation, we employ co-teaching \cite{han2018co} of a dual-branch network \cite{Li2020DivideMix, yang2023divide} to mitigate error accumulation, resulting from the confirmation bias. 
In co-teaching, due to the difference in branch parameters of the dual-branch design, error introduced by the noisy pseudolabels in one branch can be filtered out by the other branch. 
In practice, one branch conducts the clean-noisy sample separation for the other branch, and vice versa. Since each branch generates different sets of clean and noisy samples, co-teaching breaks the flow of error through the network, and thus error accumulation attenuates.
To simplify notation, the dual target branches/models $f_{t_1}$ and $f_{t_2}$ may be represented by $f_t$ in later parts of this paper. 
Both networks are trained/adapted, and the final inference can be taken from either one. 
We follow \cite{liang2022dine} to distill knowledge from the source model to the target model in a teacher-student manner. However, unlike \cite{liang2022dine}, we only have access to the hard predictions from the source model. 
Similar to \cite{liang2022dine}, the source model predictions $\hat{y}_t^i$ are updated during adaptation at certain intervals via exponential moving average between the source model predicted pseudolabels $\hat{y}_t^i$ and the target model predicted pseudolabels $y_t^i$. The process of generating $y_t^i$ is described in section \ref{ensemble}.

\subsection{Clean-noisy separation}

The predictions $\hat{y_t}$ generated by the black box source model $f_s$ are noisy and unreliable due to domain shift between $\mathcal{D}_s$ and $\mathcal{D}_t$. 
Research on learning with noisy labels shows that deep learning models tend to fit on the clean samples first, and on the noisy samples later during training. \cite{arpit2017closer,Li2020DivideMix}
We follow this insight and separate the target domain data into a clean sample set $\mathcal{X}_{tc}$ with reliable predictions, and a noisy sample set $\mathcal{X}_{tn}$ with unreliable predictions. 
In traditional noisy label settings, the noisy labels are either caused by wrong annotations from humans or from image search engines. The noise rate is, therefore, bounded. 
However, as the noisy labels in UDA are generated by the source model, the noise rate in this case is unbounded and can approach unity \cite{yi2023when}. 
We propose Jensen-Shannon distance (JSD) \cite{JSD} between the source predicted hard labels $\hat{y}_t^i$ and the target features as the criterion for clean-noisy sample separation under unbounded noise rate. JSD is calculated as, 
\begin{equation}
    JSD(\hat{y}_t^i, p_t^i) = \\
    \frac{1}{2}KL(\hat{y}_t^i, \frac{\hat{y}_t^i + p_t^i}{2}) + \frac{1}{2}KL(p_t^i, \frac{p_t^i + \hat{y}_t^i}{2})
    \label{eq:1}
\end{equation}
where, $KL(a,b)$ is the Kullback-Leibler divergence between $a$ and $b$, and $p_t^i$ is the target model output probability for target sample $x_t^i$.
Compared to cross-entropy loss, JSD is symmetric by design, and ranges between 0 and 1, thus becoming less susceptible to noise. 
When applied to the network response,
JSD produces a bimodal distribution, which is modelled by a 2-component Multivariate Gaussian Mixture Model (GMM) with equal priors.
In DA, the target model may \textit{confidently} categorize an image as the wrong class with very high prediction probability. 
Therefore, this is a poor criterion for identifying whether a sample is clean or noisy.
For the potentially unbounded pseudolabel noise rate in DABP, we take the probability of belonging to the JSD Gaussian distribution with the lower mean value as the confidence metric of being a clean sample in our clean-noisy sample separation stage.
%The samples lying on the JSD distribution with the lower mean are to be taken as the clean labeled samples. 
%Empirically, we take the probability of lying on the JSD distribution with the lower mean as a confidence metric. 
Empirically, we apply a threshold $\delta_t$ on our confidence score of belonging to the lower-mean GMM distribution to select our clean sample set $\mathcal{X}_{tc}$, at the beginning of each epoch for adaptation. 
The remaining target samples are included in the noisy label set $\mathcal{X}_{tn}$.

\subsection{Ensemble based pseudolabeling}
\label{ensemble}

\begin{figure}
\begin{minipage}[b]{1.0\linewidth}
  \centering
    \includegraphics[width=0.8\textwidth]{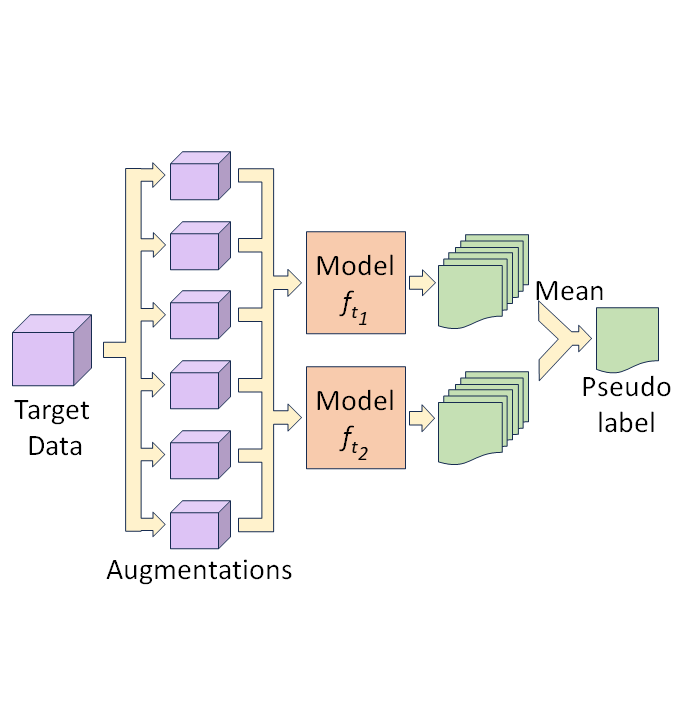}
\end{minipage}
    \caption{Ensemble-based pseudolabeling in CABB. Each sample is augmented to produce 6 different views that are fed through both branches $f_{t_1}$ and $f_{t_2}$ to create a total of 12 output predictions, which are then averaged to produce the soft pseudolabel for co-training $f_{t_1}$ and $f_{t_2}$.}
    \label{fig:ensemble_pseudolabeling}
\end{figure}

In order to produce robust target model pseudolabels $y_t^i$, we apply a series of augmentations on the target samples and produce an ensemble of output prediction probabilities from our two target models. 
We give equal weights to each output prediction and take the mean of the outputs as the soft pseudolabel as follows.

\begin{equation}
    y_t^i = \frac{1}{2M} \sum_0^{M} f_{t_1}(x_{t_m}^i)+f_{t_2}(x_{t_m}^i)
    \label{eq:pseudolabel1}
\end{equation}
where $M$ is the number of augmentations for the $i$-th target sample.
The predictions are further sharpened with a temperature factor $T (0<T<1)$ and then normalized as follows.
\begin{equation}
    y_t^i = \frac{(y_t^i)^\frac{1}{T}}{\sum_C (y_t^{iC})^\frac{1}{T}} 
    \label{eq:pseudolabel2}
\end{equation}
where $y_t^{iC}$ is the $C$-th dimensional value of the pseudolabel vector $y_t^i$.

\subsection{Curriculum guided noisy learning}

In order to mitigate early training time memorization \cite{arpit2017closer}  induced from noisy labels during the adaptation of deep models, we introduce a curriculum guided learning to train the target model on the clean samples first, and on the noisy samples later. As the adaptation/training progresses, more noisy samples are reclassified as clean samples. 

We employ separate training losses for the clean and noisy sample set. The clean set is trained with standard cross-entropy (CE) loss as follows.

\begin{equation}
    \mathcal{L}_{tc}(f_t;\mathcal{X}_{tc}) = -\mathbb{E}_{x_t^i \in \mathcal{X}_{tc}} \sum_{k=1}^{C} y_{t_k}^i log(\sigma_k(f_t(x_t^i)))
    \label{eq:L_tc}
\end{equation}
where $\sigma_k(a) = \frac{exp(a_k)}{\sum_i exp(a_i)}$ is the softmax function and $C$ is the number of classes. For the noisy set, we minimize a combination of active-passive losses \cite{ma2020normalized} constructed of normalized cross-entropy loss $\mathcal{L}_{tn_{NCE}}$ and reverse cross-entropy loss $\mathcal{L}_{tn_{RCE}}$. \cite{ma2020normalized} showed that such normalization makes a model robust to noisy data. Reverse cross-entropy loss is applied to avoid any underfitting on the noisy set. Due to the unbounded nature of noise rate in UDA and conservative clean-noisy separation criteria in CABB, we employ this particular combination of active-passive losses as our noisy set loss $\mathcal{L}_{tn}$ to make target training/adaptation robust and comprehensive on the noisy sample set. The loss function is expressed as follows.

%\begin{equation}
\begin{multline}
    \mathcal{L}_{tn_{NCE}}(f_t;\mathcal{X}_{tn}) = \\ -\mathbb{E}_{x_t^i \in \mathcal{X}_{tn}} \frac{ \sum_{k=1}^{C} y_{t_k}^i log(\sigma_k(f_t(x_t^i)))}
    {\sum_{j=1}^{C} \sum_{k=1}^{C} y_{t_j}^i log(\sigma_k(f_t(x_t^i)))}
    \label{eq:L_tn1}
\end{multline}
%\end{equation}

\begin{equation}
    \mathcal{L}_{tn_{RCE}}(f_t;\mathcal{X}_{tn}) = -\mathbb{E}_{x_t^i \in \mathcal{X}_{tn}} \sum_{k=1}^{C} \sigma_k(f_t(x_t^i)) log(y_{t_k}^i)
    \label{eq:L_tn2}
\end{equation}

\begin{equation}
    \mathcal{L}_{tn} = \mathcal{L}_{tn_{NCE}} + \beta \mathcal{L}_{tn_{RCE}}
    \label{eq:L_tn}
\end{equation}
where $\beta$ is a hyperparameter.
%\begin{equation}
%    \mathcal{L}_{tn}(f_t;\mathcal{X}_{tn}) = -\mathbb{E}_{x_t^i \in \mathcal{X}_{tn}} \sum_{k=1}^{C} || \sigma_k(f_t(x_t^i)) - y_{t_k}^i ||^2
%    \label{eq:L_tn}
%\end{equation}

To promote learning of clean samples first and to mitigate noisy label memorization, target training is done under curriculum guidance \cite{bengio2009curriculum}. 
Based on the success of the clean-noisy sample separation, the pseudolabels in the clean sample set $\mathcal{X}_{tc}$ are more likely to be correct, while those in the noisy sample set $\mathcal{X}_{tn}$ have a much higher noise rate. 
Therefore, a deep network tends to easily learn from the unambiguous $\mathcal{X}_{tc}$ set. 
We set a curriculum factor $\gamma_n$ according to the following equation.

\begin{equation}
    \gamma_n = \gamma_{n-1} (1-\alpha \epsilon^{-L_{x_n}/L_{x_{n-1}}})
    \label{eq:gamma}
\end{equation}

where, $\alpha$ is a hyperparameter and $n$ is the iteration number. 
$\gamma_{n-1}$ is the curriculum factor for the previous iteration. The ratio $L_{x_n}/L_{x_{n-1}}$ determines how much the curriculum factor decreases from iteration $n-1$ to $n$. 
If the CE loss on the clean set increases, $\gamma$ decreases by a small value to allow for further training on the clean set in the subsequent iterations. 
But if the CE loss decreases by a large margin, $\gamma$ decreases accordingly to accommodate learning from the noisy sample set in the coming iterations. 
Our curriculum guidance balances the supervised and unsupervised losses on the respective clean and noisy sets as follows.

\begin{equation}
    \mathcal{L}_{t} = \gamma_n \mathcal{L}_{tc} + (1-\gamma_n) \mathcal{L}_{tn}
    \label{eq:L_t}
\end{equation}

We adopt the formulation of information maximization (IM) loss \cite{krause2010discriminative,shi2012information,shot} from \cite{conda} to help our model produce precise predictions, while maintaining a global diversity across all classes in the output predictions. 
The IM loss is a combination of the following entropy loss $\mathcal{L}_{ent}$ and equal diversity loss $\mathcal{L}_{eqdiv}$.

\begin{equation}
%\begin{gathered}
    \mathcal{L}_{ent}(f_t;\mathcal{X}_{t}) = -\mathbb{E}_{x_t^i \in \mathcal{X}_{t}} \sum_{k=1}^{C} \sigma_k(f_t(x_t^i))log(\sigma_k(f_t(x_t^i))) \\
    \label{eq:ent}
\end{equation}
\begin{equation}
    \mathcal{L}_{eqdiv}(f_t;\mathcal{X}_{t}) = \sum_{k=1}^{C} q_k log \left ( \frac{q_k}{\hat{q}_k} \right )
    \label{eq:eqdiv}
%\end{gathered}
\end{equation}
\noindent
where $\hat{q}_k = \mathbb{E}_{x_t \in \Tilde{X}_t^*} [ \sigma(f_t(x_t)) ]$ is the mean of the softmax of the target network output response. $\mathcal{L}_{eqdiv}$ conducts KL divergence between $\hat{q}_k$ and the ideal uniform response $q_k$. 
Our curriculum guided IM loss is as follows.

\begin{equation}
    \mathcal{L}_{IM} = \mathcal{L}_{eqdiv} + (1-\gamma_n) \mathcal{L}_{ent}
    \label{eq:L_im}
\end{equation}

Minimization of entropy loss $\mathcal{L}_{ent}$ is gradually activated as the model sufficiently adapts to the clean sample. 
Such curriculum guidance ensures that the potentially erroneous predictions produced in the early stages of self-training are not accumulated. 
The $\mathcal{L}_{eqdiv}$ loss enforces diversity in the output predictions throughtout the training process. 
The overall objective function is,

\begin{equation}
    \mathcal{L}_{tot} = \mathcal{L}_{t} + \mathcal{L}_{IM}
    \label{eq:L_tot}
\end{equation}

A brief demonstration of the CABB pipeline can be found in Algorithm \ref{alg:algo}.

\begin{algorithm}[!ht]
\caption{Pseudocode for CABB}\label{alg:algo}
\DontPrintSemicolon
  \KwInput{Black-box source trained model $f_s$ and target data $x_t^i \in \mathcal{X}_{t}$}
  \KwOutput{Target adapted model $f_t$}
  \KwInit{Dual target models $f_{t_1}$ and $f_{t_2}$}
  \For{epoch = $1$ to $epoch_{total}$}
    {\While{$m \leq iter_{distill}$}{
        Distill from teacher $f_s$ to students $f_{t_1}$ and $f_{t_2}$ following \cite{liang2022dine}\;}
    Conduct clean($\mathcal{X}_{tc}$)-noisy($\mathcal{X}_{tn}$) sample separation using JSD from model $f_{t_1}$ for $f_{t_2}$ and vice-versa \;
    \For{$f_t \in f_{t_1}, f_{t_2}$}
        {\While{$n \leq iter_{adapt}$}
            {Get ensemble averaged pseudolabels $y_t^i \in \mathcal{Y}_{t}$ from equations \ref{eq:pseudolabel1} and \ref{eq:pseudolabel2} \;
            Calculate $\mathcal{L}_{tc}$ on $\mathcal{X}_{tc}$, $\mathcal{L}_{tn}$ on $\mathcal{X}_{tn}$, and $\mathcal{L}_{ent}$ and $\mathcal{L}_{eqdiv}$ on ($\mathcal{X}_{tc},\mathcal{X}_{tn}) \in \mathcal{X}_{t}$ using equations \ref{eq:L_tc}, \ref{eq:L_tn}, \ref{eq:ent}, and \ref{eq:eqdiv} respectively \;
            Calculate $\gamma_n$ using equation \ref{eq:gamma} \;
            Calculate $\mathcal{L}_{t}$ and $\mathcal{L}_{IM}$ using equations \ref{eq:L_t} and \ref{eq:L_im} \;
            Optimize $f_t$ with loss $\mathcal{L}_{tot}$ using equation \ref{eq:L_tot} \;
            }
        }
    }
%\caption{Pseudocode for CABB}
\end{algorithm}

\section{Experimental setup}

\begin{figure}
\begin{minipage}[b]{1.0\linewidth}
  \centering
    \includegraphics[width=1\textwidth]{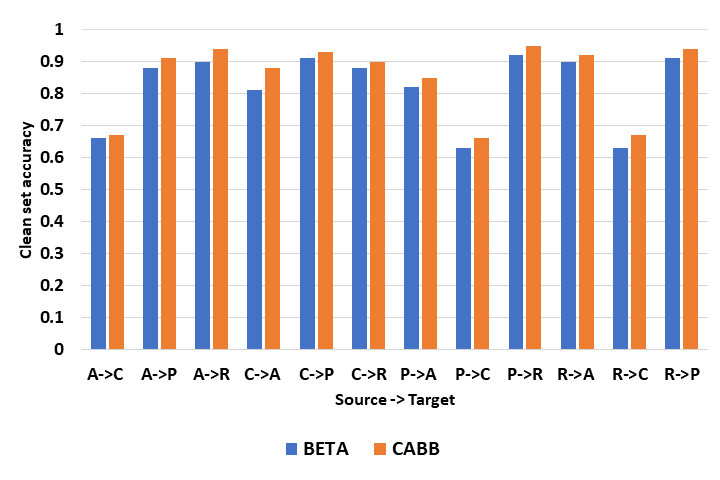}
\end{minipage}
    \caption{Accuracy on the clean sample set achieved via clean-noisy sample separation using low JSD (CABB) vs low CE (BETA), after distillation from the source teacher at the first epoch.}
    \label{fig:clean-noisy acc}
\end{figure}

\begin{table*}[htbp]
\begin{center}
\setlength\tabcolsep{5pt}
\resizebox{.6\textwidth}{!}{
\begin{tabular}{l|c|c|cccccc|c}
% \begin{tabular}{l|c|c|cccccc|c}
\toprule
Method & SF & BB & A$\rightarrow$D & A$\rightarrow$W & D$\rightarrow$A & D$\rightarrow$W & W$\rightarrow$A & W$\rightarrow$D & Mean\\
\midrule
DANN \cite{ganin2016domain} & \tikzxmark & \tikzxmark & 79.7 & 82.0 & 68.2 & 96.9 & 67.4 & 99.1 & 82.2 \\
ALDA \cite{chen2020adversarial} & \tikzxmark & \tikzxmark & 94.0 & 95.6 & 72.2 & 97.7 & 72.5 & 100.0 & 88.7 \\
GVB-GD \cite{cui2020gradually} & \tikzxmark & \tikzxmark & 95.0 & 94.8 & 73.4 & 98.7 & 73.7 & 100.0 & 89.4\\
SRDC \cite{tang2020unsupervised} & \tikzxmark & \tikzxmark & 95.8 & 95.7 & 76.7 & 99.2 & 77.1 & 100.0 & 90.9 \\
\midrule
SHOT \cite{shot} & \tikzcmark & \tikzxmark & 94.0 & 90.1 & 74.7 & 98.4 & 74.3 & 99.9 & 88.6 \\
$A^{2}$Net \cite{xia2021adaptive} & \tikzcmark & \tikzxmark & 94.5 & 94.0 & 76.7 & 99.2 & 76.1 & 100 & 90.1 \\
SFDA-DE \cite{ding2022source} & \tikzcmark & \tikzxmark & 96.0 & 94.2 & 76.6 & 98.5 & 75.5 & 99.8 & 90.1 \\
\midrule
LNL-OT \cite{YM.2020Self-labelling} & \tikzcmark & \tikzcmark & 88.8 & 85.5 & 64.6 & 95.1 & 66.7 & 98.7 & 83.2 \\
LNL-KL \cite{zhang2021unsupervised} & \tikzcmark & \tikzcmark & 89.4 & 86.8 & 65.1 & 94.8 & 67.1 & 98.7 & 83.6 \\
HD-SHOT \cite{liang2021source} & \tikzcmark & \tikzcmark & 86.5 & 83.1 & 66.1 & 95.1 & 68.9 & 98.1 & 83.0 \\
SD-SHOT \cite{liang2021source} & \tikzcmark & \tikzcmark & 89.2 & 83.7 & 67.9 & 95.3 & 71.1 & 97.1 & 84.1 \\ 
DINE \cite{liang2022dine} & \tikzcmark & \tikzcmark &  91.6 & 86.8 & 72.2 & 96.2 & 73.3 & 98.6 & 86.4 \\
BETA \cite{yang2023divide} & \tikzcmark & \tikzcmark & 93.6 & 88.3 & \textbf{76.1} & 95.5 & \textbf{76.5} & 99.0 & 88.2 \\
\midrule
\rowcolor{maroon!10}
CABB (Ours) & \tikzcmark & \tikzcmark &  \textbf{94.0} & \textbf{88.6} & \textbf{76.0} & \textbf{97.9} & 76.0 & \textbf{99.6} & \textbf{88.7} \\

\bottomrule
\end{tabular}}
\end{center}
\caption{Mean accuracy on the Office31. 'SF' refers to source-free and 'BB' means black-box. The top performing results among the DABP methods are in bold letters.}
\label{res:officehome}
\label{tab:office31res}
\end{table*}

\subsection{Datasets}
We evaluate CABB on three popular domain adaptation datasets viz. Office-31 \cite{saenko2010adapting}, Office-Home \cite{venkateswara2017deep}, and VisDA-C \cite{peng2018visda}.
\textbf{Office-31} is a small-scale DA dataset consisting of images of 31 classes of common objects found in an office across 3 domains viz. Amazon (A), Webcam (W), and DSLR (D). 
\textbf{Office-Home} is a medium-sized DA dataset consisting of 4 domains viz. Art (A), Clipart (C), Product (P), and Real-World (R). The dataset contains images of 65 classes of items found in office and home environments.
\textbf{VisDA-C} is a large-scale dataset consisting of 12 classes of objects across 2 domains: Synthetic (S) and Real (R). The 152K synthetic images are generated by 3D rendering and taken as the source domain. The 55K real samples are taken from MS COCO dataset \cite{lin2014microsoft} and taken as the target domain.

\subsection{Implementation details}
We follow the same protocol in \cite{liang2022dine, yang2023divide} for source training to ensure fairness for comparison. 
Our target models are initialized with ImageNet pretrained weights, since source model parameters are inaccessible. 
For Office-31 and Office-Home, we use ResNet50, and for VisDA-C we use ResNet101 as the backbone \cite{he2016deep}, on top of which we attach an MLP-based classifier, similar to \cite{liang2022dine, yang2023divide}. 
The target models are trained with SGD optimizer with $0.9$ momentum and weight decay $1e^{-3}$. 
The learning rate for the backbone is set to $1e^{-3}$, while that of the classifier is set to $1e^{-2}$. 
$\alpha$ in the curriculum factor is set to $2e^{-3}$ for Office-31, and $2e^{-4}$ for Office-Home and VisDA-C, depending on the size of the dataset.
The model is adapted for 50 epochs for Office-31 and Office-Home datasets, and for 5 epochs for the VisDA-C dataset.
Temperature sharpening factor $T$ is set to $0.5$. We implement our method using the PyTorch library on an NVIDIA-A100 GPU.

\section{Results}
\subsection{Overall evaluation}

\begin{table*}[htbp]
\begin{center}
\resizebox{\textwidth}{!}{
\begin{tabular}{l|c|c|cccccccccccc|c}
% \begin{tabular}{l|c|c|cccccccccccc|c}
\toprule
Method & SF & BB & A$\rightarrow$C & A$\rightarrow$P & A$\rightarrow$R & C$\rightarrow$A & C$\rightarrow$P & C$\rightarrow$R & P$\rightarrow$A & P$\rightarrow$C & P$\rightarrow$R & R$\rightarrow$A & R$\rightarrow$C & R$\rightarrow$P & Mean\\
\midrule

DANN \cite{ganin2016domain} & \tikzxmark & \tikzxmark & 45.6 & 59.3 & 70.1 & 47.0 & 58.5 & 60.9 & 46.1 & 43.7 & 68.5 & 63.2 & 51.8 & 76.8 & 57.6 \\
ALDA \cite{chen2020adversarial} & \tikzxmark & \tikzxmark & 53.7 & 70.1 & 76.4 & 60.2 & 72.6 & 71.5 & 56.8 & 51.9 & 77.1 & 70.2 & 56.3 & 82.1 & 66.6 \\
GVB-GD \cite{cui2020gradually} & \tikzxmark & \tikzxmark & 57.0 & 74.7 & 79.8 & 64.6 & 74.1 & 74.6 & 65.2 & 55.1 & 81.0 & 74.6 & 59.7 & 84.3 & 70.4 \\
SRDC \cite{tang2020unsupervised} & \tikzxmark & \tikzxmark & 52.3 & 76.3 & 81.0 & 69.5 & 76.2 & 78.0 & 68.7 & 53.8 & 81.7 & 76.3 & 57.1 & 85.0 & 71.3 \\
FixBi \cite{na2021fixbi} & \tikzxmark & \tikzxmark & 58.1 & 77.3 & 80.4 & 67.7 & 79.5 & 78.1 & 65.8 & 57.9 & 81.7 & 76.4 & 62.9 & 86.7 & 72.7 \\
\midrule
G-SFDA \cite{yang2021generalized} & \tikzcmark & \tikzxmark & 57.9 & 78.6 & 81.0 & 66.7 & 77.2 & 77.2 & 65.6 & 56.0 & 82.2 & 72.0 & 57.8 & 83.4 & 71.3 \\
SHOT \cite{shot} & \tikzcmark & \tikzxmark & 57.1 & 78.1 & 81.5 & 68.0 & 78.2 & 78.1 & 67.4 & 54.9 & 82.2 & 73.3 & 58.8 & 84.3 & 71.8 \\
HCL \cite{huang2021model} & \tikzcmark & \tikzxmark & 64.0 & 78.6 & 82.4 & 64.5 & 73.1 & 80.1 & 64.8 & 59.8 & 75.3 & 78.1 & 69.3 & 81.5 & 72.6 \\
$A^{2}$Net \cite{xia2021adaptive} & \tikzcmark & \tikzxmark & 58.4 & 79.0 & 82.4 & 67.5 & 79.3 & 78.9 & 68.0 & 56.2 & 82.9 & 74.1 & 60.5 & 85.0 & 72.8 \\
SFDA-DE \cite{ding2022source} & \tikzcmark & \tikzxmark & 59.7 & 79.5 & 82.4 & 69.7 & 78.6 & 79.2 & 66.1 & 57.2 & 82.6 & 73.9 & 60.8 & 85.5 & 72.9 \\

\midrule
LNL-OT \cite{YM.2020Self-labelling} & \tikzcmark & \tikzcmark & 49.1 & 71.7 & 77.3 & 60.2 & 68.7 & 73.1 & 57.0 & 46.5 & 76.8 & 67.1 & 52.3 & 79.5 & 64.9 \\ 
LNL-KL \cite{zhang2021unsupervised} & \tikzcmark & \tikzcmark & 49.0 & 71.5 & 77.1 & 59.0 & 68.7 & 72.9 & 56.4 & 46.9 & 76.6 & 66.2 & 52.3 & 79.1 & 64.6 \\
HD-SHOT \cite{liang2021source} & \tikzcmark & \tikzcmark & 48.6 & 72.8 & 77.0 & 60.7 & 70.0 & 73.2 & 56.6 & 47.0 & 76.7 & 67.5 & 52.6 & 80.2 & 65.3 \\ 
SD-SHOT \cite{liang2021source} & \tikzcmark & \tikzcmark & 50.1 & 75.0 & 78.8 & 63.2 & 72.9 & 76.4 & 60.0 & 48.0 & 79.4 & 69.2 & 54.2 & 81.6 & 67.4 \\
DINE \cite{liang2022dine} & \tikzcmark & \tikzcmark &  52.2 & 78.4 & 81.3 & 65.3 & 76.6 & 78.7 & 62.7 & 49.6 & 82.2 & 69.8 & 55.8 & 84.2 & 69.7  \\
BETA \cite{yang2023divide} & \tikzcmark & \tikzcmark & 57.2 & 78.5 & \textbf{82.1} & \textbf{68.0} & 78.6 & \textbf{79.7} & 67.5 & 56.0 & \textbf{83.0} & 71.9 & 58.9 & 84.2 & 72.1 \\
\midrule
\rowcolor{maroon!10}
CABB (Ours) & \tikzcmark & \tikzcmark &  \textbf{57.4} & \textbf{79.5} & \textbf{82.0} & \textbf{68.1} & \textbf{79.3} & 78.8 & \textbf{68.2} & \textbf{57.9} & 82.7 & \textbf{73.6} & \textbf{60.0} & \textbf{86.4} & \textbf{72.8} \\
\bottomrule
\end{tabular}}
\end{center}
\caption{Mean accuracy on the Office-Home dataset. 'SF' refers to source-free and 'BB' means black-box. The top  performing results among the DABP methods are in bold letters.}
\label{res:officehomeres}
\end{table*}

\begin{table*}[htbp]
\begin{center}
\resizebox{\textwidth}{!}{
\begin{tabular}{l|c|c|cccccccccccc|c}
% \begin{tabular}{l|c|c|cccccccccccc|c}
\toprule
Method & SF & BB & plane & bcycl & bus & car & horse & knife & mcycle & person & plant & sktbrd & train & truck & Per-class \\
\midrule
DANN \cite{ganin2016domain} & \tikzxmark & \tikzxmark & 81.9 & 77.7 & 82.8 & 44.3 & 81.2 & 29.5 & 65.2 & 28.6 & 51.9 & 54.6 & 82.8 & 7.8 & 57.6 \\
ALDA \cite{chen2020adversarial} & \tikzxmark & \tikzxmark & 93.8 & 74.1 & 82.4 & 69.4 & 90.6 & 87.2 & 89.0 & 67.6 & 93.4 & 76.1 & 87.7 & 22.2 & 77.8 \\
\midrule
SHOT \cite{shot} & \tikzcmark & \tikzxmark &94.3 & 88.5 & 80.1 & 57.3 & 93.1 & 94.9 & 80.7 & 80.3 & 91.5 & 89.1 & 86.3 & 58.2 & 82.9 \\
$A^{2}$Net \cite{xia2021adaptive} & \tikzcmark & \tikzxmark & 94.0 & 87.8 & 85.6 & 66.8 & 93.7 & 95.1 & 85.8 & 81.2 & 91.6 & 88.2 & 86.5 & 56.0 & 84.3 \\
SFDA-DE \cite{ding2022source} & \tikzcmark & \tikzxmark & 95.3 & 91.2 & 77.5 & 72.1 & 95.7 & 97.8 & 85.5 & 86.1 & 95.5 & 93.0 & 86.3 & 61.6 & 86.5 \\
\midrule
LNL-OT \cite{YM.2020Self-labelling} & \tikzcmark & \tikzcmark & 82.6 & 84.1 & 76.2 & 44.8 & 90.8 & 39.1 & 76.7 & 72.0 & 82.6 & 81.2 & 82.7 & 50.6 & 72.0 \\ 
LNL-KL \cite{zhang2021unsupervised} & \tikzcmark & \tikzcmark & 82.7 & 83.4 & 76.7 & 44.9 & 90.9 & 38.5 & 78.4 & 71.6 & 82.4 & 80.3 & 82.9 & 50.4 & 71.9 \\
HD-SHOT \cite{liang2021source} & \tikzcmark & \tikzcmark & 75.8 & 85.8 & 78.0 & 43.1 & 92.0 & 41.0 & 79.9 & 78.1 & 84.2 & 86.4 & 81.0 & 65.5 & 74.2 \\
SD-SHOT \cite{liang2021source} & \tikzcmark & \tikzcmark & 79.1 & 85.8 & 77.2 & 43.4 & 91.6 & 41.0 & 80.0 & 78.3 & 84.7 & 86.8 & 81.1 & 65.1 & 74.5 \\
DINE \cite{liang2022dine} & \tikzcmark & \tikzcmark &  81.4 & 86.7 & 77.9 & 55.1 & 92.2 & 34.6 & 80.8 & 79.9 & 87.3 & 87.9 & 84.3 & 58.7 & 75.6 \\
BETA \cite{yang2023divide} & \tikzcmark & \tikzcmark & \textbf{96.2} & 83.9 & 82.3 & 71.0 & \textbf{95.3} & 73.1 & \textbf{88.4} & 80.6 & \textbf{95.5} & 90.9 & \textbf{88.3} & 45.1 & 82.6 \\
\midrule
\rowcolor{maroon!10}
CABB (Ours) & \tikzcmark & \tikzcmark & 95.1 & \textbf{87.0} & \textbf{82.6} & \textbf{71.5} & 94.5 & \textbf{89.7} & 87.5 & \textbf{81.5} & 93.8 & \textbf{92.4} & 87.3 & \textbf{55.5} & \textbf{84.9} \\
\bottomrule
\end{tabular}}
\end{center}
\caption{Mean per-class accuracy on the VisDA-C dataset. 'SF' refers to source-free and 'BB' means black-box. The top  performing results among the DABP methods are in bold letters.}
\label{res:visdares}
\end{table*}

\begin{table*}[htbp]
\begin{center}
\resizebox{\textwidth}{!}{
\begin{tabular}{L|L|L|cccccccccccc|c}
\toprule
Curriculum & $\mathcal{L}_{tn}$ & $\mathcal{L}_{ent}$ & plane & bcycl & bus & car & horse & knife & mcycle & person & plant & sktbrd & train & truck & Per-class \\
\midrule
\centering
\tikzxmark & \tikzxmark & \tikzxmark & 98.0 & 93.1 & 79.1 & 41.8 & 97.1 & 81.6 & 79.5 & 79.9 & 93.3 & 91.1 & 90.3 & 49.5 & 81.2 \\
\tikzxmark & \tikzxmark & \tikzcmark & 98.2 & 89.2 & 82.2 & 58.1 & 97.2 & 83.5 & 84.3 & 71.3 & 95.8 & 92.2 & 90.4 & 18.1 & 80.0 \\
\tikzxmark & \tikzcmark & \tikzcmark & 97.1 & 82.3 & 85.0 & 79.1 & 91.7 & 93.2 & 89.0 & 77.7 & 94.4 & 92.5 & 83.9 & 1.2 & 80.6 \\
\midrule
\tikzcmark & \tikzxmark & \tikzcmark & 97.3 & 89.9 & 78.3 & 60.1 & 96.4 & 76.1 & 80.2 & 77.3 & 93.5 & 90.0 & 88.7 & 52.8 & 81.7 \\
\tikzcmark & \tikzcmark & \tikzxmark & 95.2 & 85.9 & 83.5 & 68.9 & 93.8 & 88.6 & 83.6 & 80.7 & 95.1 & 92.0 & 86.0 & 56.7 & 84.2 \\
%\tikzxmark & \tikzxmark & \tikzxmark & 98.0 & 93.1 & 79.1 & 41.8 & 97.1 & 81.6 & 79.5 & 79.9 & 93.3 & 91.1 & 90.3 & 49.5 & 81.2 \\
%\tikzxmark & \tikzcmark & \tikzxmark & 97.4 & 86.7 & 81.0 & 72.7 & 95.3 & 93.7 & 86.4 & 80.7 & 95.0 & 92.3 & 89.2 & 38.4 & 84.1 \\
%\tikzxmark & \tikzcmark & \tikzcmark & 97.1 & 82.3 & 85.0 & 79.1 & 91.7 & 93.2 & 89.0 & 77.7 & 94.4 & 92.5 & 83.9 & 1.2 & 80.6 \\
% \tikzxmark & \tikzxmark & \tikzcmark & 98.2 & 89.2 & 82.2 & 58.1 & 97.2 & 83.5 & 84.3 & 71.3 & 95.8 & 92.2 & 90.4 & 18.1 & 80.0 \\
\tikzcmark & \tikzcmark & \tikzcmark & 95.1 & 87.0 & 82.6 & 71.5 & 94.5 & 89.7 & 87.5 & 81.5 & 93.8 & 92.4 & 87.3 & 55.5 & 84.9 \\

\bottomrule
\end{tabular}}
\end{center}
\caption{Performance evaluation of curriculum adaptation involving different parts of CABB on the VisDA-C dataset. The 'tick' marks mean the part is present in the model, and the 'cross' mark means that part is absent. When curriculum is absent and $\mathcal{L}_{tn}$ is present, $\gamma_n$ is set to $0.5$.}
\label{res:visdaablation}
\end{table*}

%We compare CABB against state-of-the-art black-box DA models. 
Liang \textit{et al.} \cite{liang2022dine} pioneered this area and formulated the problem statement.
They also presented a number of baselines for comparison. 
Among them \textbf{NLL-KD} and \textbf{NLL-OT} are inspired by noisy label learning and utilize KL divergence and optimal transport respectively for refining pseudolabels.
\textbf{HD-SHOT} and \textbf{SD-SHOT} are based on the SHOT \cite{shot} model and treat the source model predictions as hard labels and soft labels, respectively. 
In addition to these baselines, we compare CABB against state-of-the-art black-box DA models \textbf{DINE} \cite{liang2022dine} and \textbf{BETA} \cite{yang2023divide}. 
We further compare against a number of standard DA methods, such as \textbf{DANN} \cite{ganin2016domain}, \textbf{ALDA} \cite{chen2020adversarial}, \textbf{GVB-GD} \cite{cui2020gradually}, \textbf{SRDC} \cite{tang2020unsupervised}, \textbf{SHOT} \cite{shot}, \textbf{A$^2$-Net} \cite{xia2021adaptive}, \textbf{SFDA-DE} \cite{ding2022source} etc.

In Figure \ref{fig:clean-noisy acc}, we present the accuracy of the clean sample set after clean-noisy sample separation for the first epoch after distillation from the source teacher model to the target student model. 
We can see that our choice of low JSD separation criterion in CABB consistently outperforms the low CE loss criterion used in BETA by 1\%-7\% across all 12 source-target domain pairs for Office-Home dataset. 

The classification accuracies after adaptation across the 6 domain pairs for Office-31 dataset are shown in Table \ref{tab:office31res}. 
CABB outperforms BETA and DINE on average by 0.5\% and 2.3\%, respectively. 
While CABB beats DINE across all the domain pairs, it only underperforms BETA for \textbf{Webcam-Amazon} adaptation by 0.5\%. 
Overall, CABB is on-par with \textit{white-box source-free} model SHOT and \textit{non-source-free} model ALDA.

The results for Office-Home dataset are presented in Table \ref{res:officehomeres}. 
CABB outperforms BETA and DINE by 0.7\% and 3.1\%, respectively. 
Moreover, CABB outperforms several standard \textit{non-source-free} DA methods such as SRDC and FixBi, and is either better than, or on par with existing state-of-the-art \textit{white-box source-free} DA models like HCL, $A^2$Net, and SFDA-DE.

A comparative evaluation of CABB against other state-of-the-art DA methods and DABP baselines on the VisDA-C dataset is shown in Table \ref{res:visdares}.
CABB surpasses both DINE and BETA by $9.3\%$ and $2.3\%$, respectively in terms of mean-per-class accuracy.
CABB beats BETA in the most challenging catergory \textit{truck} by 10.4\%. 
CABB also outperforms \textit{white-box source-free} models SHOT and $A^2$Net comfortably. 

\subsection{Ablation study}
A detailed ablation study on the efficacy of our curriculum adaptation method is given in Table \ref{res:visdaablation}. The impact of curriculum on the noisy set loss $\mathcal{L}_{tn}$ and entropy loss $\mathcal{L}_{ent}$ is shown, as curriculum is applied to these two components.
In this table, in the absence of curriculum adaptation, $\gamma_n$ is set to $0.5$. In row 2, $\mathcal{L}_{tn}$ is set to $0$.

The results clearly indicate the benefit of a guided adaptation framework that progressively learns from the clean samples first and the noisy samples later. We see in the first three rows in Table \ref{res:visdaablation} that without curriculum guidance, adaptation performance suffers significantly.
In the absence of curriculum guidance, we see that leaving out learning from the noisy samples during the adaptation process is better than adapting to the noisy samples with $\mathcal{L}_{tn}$ loss, and further enforcing the wrong predictions with $\mathcal{L}_{ent}$ loss.
The drawback of blindly adapting to noisy samples becomes evident in the second and third rows, particularly in the most challenging \textit{truck} class. 
By adapting to unrefined noisy samples from the beginning, the model performance drastically deteriorates and accuracy on \textit{truck} can fall to as low as $~1\%$.

The results in the 4th through 6th rows in Table \ref{res:visdaablation} show the necessity for curriculum guidance during adaptation. 
In the presence of curriculum learning, CABB outperforms existing state-of-the-art DABP methods. 
Curriculum guidance progressively refines the noisy sample pseudolabels.
While enforcing the refined predictions by minimizing the $\mathcal{L}_{ent}$ loss produces improved results, learning from the noisy pseudolabels by minimizing the $\mathcal{L}_{tn}$ loss significantly boosts the model performance. 
Minimizing losses $\mathcal{L}_{tn}$ and $\mathcal{L}_{ent}$ on the refined pseudolabels together produce the strongest results. 

\section{Conclusion}
In this paper we present a curriculum guided self-training based domain adaptation method called CABB to adapt a black-box source model/predictor to the target domain. 
Without access to the source data or the source model parameters during adaptation, we draw inspiration from noisy label learning algorithms. 
We employ a co-training scheme and propose to use Jensen-Shannon distance or JSD as the criterion to filter clean and reliable samples from noisy and unreliable samples. 
JSD calculated between the source model predicted pseudolabels and target model predictions is modelled using a mixture of Gaussian distributions. 
The samples with high probability of lying on the distribution with the lower mean JSD are taken as clean samples, and the target model is trained under a curriculum schedule first on the clean samples and progressively on the noisy samples. 
The dual-branch design of CABB also allows robust ensemble-based pseudolabeling.  
CABB consistently outperforms existing black-box domain adaptation models on three popular domain adaptation benchmarks, and is on par with other white-box source free models.

%\section{File Naming}

%Figures (line artwork or photographs) should be named starting with the first 5 letters of the author’s last name. The next characters in the filename should be the number that represents the sequential location of this image in your article. For example, in author ``Anderson's'' paper, the first three figures would be named ander1.tif, ander2.tif, and ander3.ps.

%Tables should contain only the body of the table (not the caption) and should be named similarly to figures, except that `.t' is inserted in-between the author's name and the table number. For example, author Anderson's first three tables would be named ander.t1.tif, ander.t2.ps, ander.t3.eps.

%Author photographs should be named using the first five characters of the pictured author's last name. For example, four author photographs for a paper may be named: oppen.ps, moshc.tif, chen.eps, and duran.pdf.

%If two authors or more have the same last name, their first initial(s) can be substituted for the fifth, fourth, third... letters of their surname until the degree where there is differentiation. For example, two authors Michael and Monica Oppenheimer's photos would be named oppmi.tif, and oppmo.eps.

\section*{Acknowledgment} 
The authors would like to thank Nazmul Karim for his valuable suggestions and insights regarding noisy label learning. The authors would also like to thank RIT Research Computing for making computing resources available for experimentation.

%The preferred spelling of the word ``acknowledgment'' in American English is without an ``e'' after the ``g.'' Use the singular heading even if you have many acknowledgments. Avoid expressions such as ``One of us (S.B.A.) would like to thank ... .'' Instead, write ``F. A. Author thanks ... .'' In most cases, sponsor and financial support acknowledgments are placed in the unnumbered footnote on the first page, not here.

%\section*{References}

\bibliographystyle{IEEEtran}
\bibliography{TAI_template}

\begin{thebibliography}{10}
\providecommand{\url}[1]{#1}
\csname url@samestyle\endcsname
\providecommand{\newblock}{\relax}
\providecommand{\bibinfo}[2]{#2}
\providecommand{\BIBentrySTDinterwordspacing}{\spaceskip=0pt\relax}
\providecommand{\BIBentryALTinterwordstretchfactor}{4}
\providecommand{\BIBentryALTinterwordspacing}{\spaceskip=\fontdimen2\font plus
\BIBentryALTinterwordstretchfactor\fontdimen3\font minus
  \fontdimen4\font\relax}
\providecommand{\BIBforeignlanguage}[2]{{%
\expandafter\ifx\csname l@#1\endcsname\relax
\typeout{** WARNING: IEEEtran.bst: No hyphenation pattern has been}%
\typeout{** loaded for the language `#1'. Using the pattern for}%
\typeout{** the default language instead.}%
\else
\language=\csname l@#1\endcsname
\fi
#2}}
\providecommand{\BIBdecl}{\relax}
\BIBdecl

\bibitem{ganin2016domain}
Y.~Ganin, E.~Ustinova, H.~Ajakan, P.~Germain, H.~Larochelle, F.~Laviolette,
  M.~Marchand, and V.~Lempitsky, ``Domain-adversarial training of neural
  networks,'' \emph{The journal of machine learning research}, vol.~17, no.~1,
  pp. 2096--2030, 2016.

\bibitem{hoffman2018cycada}
J.~Hoffman, E.~Tzeng, T.~Park, J.-Y. Zhu, P.~Isola, K.~Saenko, A.~Efros, and
  T.~Darrell, ``Cycada: Cycle-consistent adversarial domain adaptation,'' in
  \emph{International conference on machine learning}.\hskip 1em plus 0.5em
  minus 0.4em\relax Pmlr, 2018, pp. 1989--1998.

\bibitem{torralba2011unbiased}
A.~Torralba and A.~A. Efros, ``Unbiased look at dataset bias,'' in \emph{CVPR
  2011}.\hskip 1em plus 0.5em minus 0.4em\relax IEEE, 2011, pp. 1521--1528.

\bibitem{tzeng2014deep}
E.~Tzeng, J.~Hoffman, N.~Zhang, K.~Saenko, and T.~Darrell, ``Deep domain
  confusion: Maximizing for domain invariance,'' \emph{arXiv preprint
  arXiv:1412.3474}, 2014.

\bibitem{zellingercentral}
W.~Zellinger, T.~Grubinger, E.~Lughofer, T.~Natschl{\"a}ger, and
  S.~Saminger-Platz, ``Central moment discrepancy (cmd) for domain-invariant
  representation learning,'' in \emph{International Conference on Learning
  Representations}.

\bibitem{shot}
J.~Liang, D.~Hu, and J.~Feng, ``Do we really need to access the source data?
  source hypothesis transfer for unsupervised domain adaptation,'' in
  \emph{International Conference on Machine Learning}.\hskip 1em plus 0.5em
  minus 0.4em\relax PMLR, 2020, pp. 6028--6039.

\bibitem{yang2021generalized}
S.~Yang, Y.~Wang, J.~Van De~Weijer, L.~Herranz, and S.~Jui, ``Generalized
  source-free domain adaptation,'' in \emph{Proceedings of the IEEE/CVF
  International Conference on Computer Vision}, 2021, pp. 8978--8987.

\bibitem{liang2022dine}
J.~Liang, D.~Hu, J.~Feng, and R.~He, ``Dine: Domain adaptation from single and
  multiple black-box predictors,'' in \emph{Proceedings of the IEEE/CVF
  Conference on Computer Vision and Pattern Recognition}, 2022.

\bibitem{yang2023divide}
\BIBentryALTinterwordspacing
J.~Yang, X.~Peng, K.~Wang, Z.~Zhu, J.~Feng, L.~Xie, and Y.~You, ``Divide to
  adapt: Mitigating confirmation bias for domain adaptation of black-box
  predictors,'' in \emph{The Eleventh International Conference on Learning
  Representations}, 2023. [Online]. Available:
  \url{https://openreview.net/forum?id=hVrXUps3LFA}
\BIBentrySTDinterwordspacing

\bibitem{Li2020DivideMix}
\BIBentryALTinterwordspacing
J.~Li, R.~Socher, and S.~C. Hoi, ``Dividemix: Learning with noisy labels as
  semi-supervised learning,'' in \emph{International Conference on Learning
  Representations}, 2020. [Online]. Available:
  \url{https://openreview.net/forum?id=HJgExaVtwr}
\BIBentrySTDinterwordspacing

\bibitem{han2018co}
B.~Han, Q.~Yao, X.~Yu, G.~Niu, M.~Xu, W.~Hu, I.~Tsang, and M.~Sugiyama,
  ``Co-teaching: Robust training of deep neural networks with extremely noisy
  labels,'' \emph{Advances in neural information processing systems}, vol.~31,
  2018.

\bibitem{conda}
A.~M.~N. Taufique, C.~S. Jahan, and A.~Savakis, ``Continual unsupervised domain
  adaptation in data-constrained environments,'' \emph{IEEE Transactions on
  Artificial Intelligence}, 2023.

\bibitem{long2015learning}
M.~Long, Y.~Cao, J.~Wang, and M.~Jordan, ``Learning transferable features with
  deep adaptation networks,'' in \emph{International conference on machine
  learning}.\hskip 1em plus 0.5em minus 0.4em\relax PMLR, 2015, pp. 97--105.

\bibitem{sun2016deep}
B.~Sun and K.~Saenko, ``Deep coral: Correlation alignment for deep domain
  adaptation,'' in \emph{Computer Vision--ECCV 2016 Workshops: Amsterdam, The
  Netherlands, October 8-10 and 15-16, 2016, Proceedings, Part III 14}.\hskip
  1em plus 0.5em minus 0.4em\relax Springer, 2016, pp. 443--450.

\bibitem{long2017deep}
M.~Long, H.~Zhu, J.~Wang, and M.~I. Jordan, ``Deep transfer learning with joint
  adaptation networks,'' in \emph{International conference on machine
  learning}.\hskip 1em plus 0.5em minus 0.4em\relax PMLR, 2017, pp. 2208--2217.

\bibitem{pei2018multi}
Z.~Pei, Z.~Cao, M.~Long, and J.~Wang, ``Multi-adversarial domain adaptation,''
  in \emph{Proceedings of the AAAI conference on artificial intelligence},
  vol.~32, no.~1, 2018.

\bibitem{tzeng2017adversarial}
E.~Tzeng, J.~Hoffman, K.~Saenko, and T.~Darrell, ``Adversarial discriminative
  domain adaptation,'' in \emph{Proceedings of the IEEE conference on computer
  vision and pattern recognition}, 2017, pp. 7167--7176.

\bibitem{pmlr-v80-hoffman18a}
\BIBentryALTinterwordspacing
J.~Hoffman, E.~Tzeng, T.~Park, J.-Y. Zhu, P.~Isola, K.~Saenko, A.~Efros, and
  T.~Darrell, ``Cycada: Cycle-consistent adversarial domain adaptation,'' in
  \emph{Proceedings of the 35th International Conference on Machine Learning},
  ser. Proceedings of Machine Learning Research, J.~Dy and A.~Krause, Eds.,
  vol.~80.\hskip 1em plus 0.5em minus 0.4em\relax PMLR, 10--15 Jul 2018, pp.
  1989--1998. [Online]. Available:
  \url{https://proceedings.mlr.press/v80/hoffman18a.html}
\BIBentrySTDinterwordspacing

\bibitem{zhu2017unpaired}
J.-Y. Zhu, T.~Park, P.~Isola, and A.~A. Efros, ``Unpaired image-to-image
  translation using cycle-consistent adversarial networks,'' in
  \emph{Proceedings of the IEEE international conference on computer vision},
  2017, pp. 2223--2232.

\bibitem{li2019joint}
S.~Li, C.~H. Liu, B.~Xie, L.~Su, Z.~Ding, and G.~Huang, ``Joint adversarial
  domain adaptation,'' in \emph{Proceedings of the 27th ACM International
  Conference on Multimedia}, 2019, pp. 729--737.

\bibitem{pan2019transferrable}
Y.~Pan, T.~Yao, Y.~Li, Y.~Wang, C.-W. Ngo, and T.~Mei, ``Transferrable
  prototypical networks for unsupervised domain adaptation,'' in
  \emph{Proceedings of the IEEE/CVF conference on computer vision and pattern
  recognition}, 2019, pp. 2239--2247.

\bibitem{tang2020unsupervised}
H.~Tang, K.~Chen, and K.~Jia, ``Unsupervised domain adaptation via structurally
  regularized deep clustering,'' in \emph{Proceedings of the IEEE/CVF
  conference on computer vision and pattern recognition}, 2020, pp. 8725--8735.

\bibitem{chen2020graph}
L.~Chen, Z.~Gan, Y.~Cheng, L.~Li, L.~Carin, and J.~Liu, ``Graph optimal
  transport for cross-domain alignment,'' in \emph{International Conference on
  Machine Learning}.\hskip 1em plus 0.5em minus 0.4em\relax PMLR, 2020, pp.
  1542--1553.

\bibitem{xu2020reliable}
R.~Xu, P.~Liu, L.~Wang, C.~Chen, and J.~Wang, ``Reliable weighted optimal
  transport for unsupervised domain adaptation,'' in \emph{Proceedings of the
  IEEE/CVF conference on computer vision and pattern recognition}, 2020, pp.
  4394--4403.

\bibitem{chidlovskii2016domain}
B.~Chidlovskii, S.~Clinchant, and G.~Csurka, ``Domain adaptation in the absence
  of source domain data,'' in \emph{Proceedings of the 22nd ACM SIGKDD
  International Conference on Knowledge Discovery and Data Mining}, 2016, pp.
  451--460.

\bibitem{liang2019distant}
J.~Liang, R.~He, Z.~Sun, and T.~Tan, ``Distant supervised centroid shift: A
  simple and efficient approach to visual domain adaptation,'' in
  \emph{Proceedings of the IEEE/CVF Conference on Computer Vision and Pattern
  Recognition}, 2019, pp. 2975--2984.

\bibitem{krause2010discriminative}
A.~Krause, P.~Perona, and R.~Gomes, ``Discriminative clustering by regularized
  information maximization,'' \emph{Advances in neural information processing
  systems}, vol.~23, 2010.

\bibitem{shi2012information}
Y.~Shi and F.~Sha, ``Information-theoretical learning of discriminative
  clusters for unsupervised domain adaptation,'' in \emph{Proceedings of the
  29th International Coference on International Conference on Machine
  Learning}, 2012, pp. 1275--1282.

\bibitem{hu2017learning}
W.~Hu, T.~Miyato, S.~Tokui, E.~Matsumoto, and M.~Sugiyama, ``Learning discrete
  representations via information maximizing self-augmented training,'' in
  \emph{International conference on machine learning}.\hskip 1em plus 0.5em
  minus 0.4em\relax PMLR, 2017, pp. 1558--1567.

\bibitem{ding2022source}
N.~Ding, Y.~Xu, Y.~Tang, C.~Xu, Y.~Wang, and D.~Tao, ``Source-free domain
  adaptation via distribution estimation,'' in \emph{Proceedings of the
  IEEE/CVF Conference on Computer Vision and Pattern Recognition}, 2022, pp.
  7212--7222.

\bibitem{zhang2017understanding}
\BIBentryALTinterwordspacing
C.~Zhang, S.~Bengio, M.~Hardt, B.~Recht, and O.~Vinyals, ``Understanding deep
  learning requires rethinking generalization,'' in \emph{International
  Conference on Learning Representations}, 2017. [Online]. Available:
  \url{https://openreview.net/forum?id=Sy8gdB9xx}
\BIBentrySTDinterwordspacing

\bibitem{arpit2017closer}
D.~Arpit, S.~Jastrzkebski, N.~Ballas, D.~Krueger, E.~Bengio, M.~S. Kanwal,
  T.~Maharaj, A.~Fischer, A.~Courville, Y.~Bengio \emph{et~al.}, ``A closer
  look at memorization in deep networks,'' in \emph{International conference on
  machine learning}.\hskip 1em plus 0.5em minus 0.4em\relax PMLR, 2017, pp.
  233--242.

\bibitem{yi2023when}
\BIBentryALTinterwordspacing
L.~Yi, G.~Xu, P.~Xu, J.~Li, R.~Pu, C.~Ling, I.~McLeod, and B.~Wang, ``When
  source-free domain adaptation meets learning with noisy labels,'' in
  \emph{The Eleventh International Conference on Learning Representations},
  2023. [Online]. Available: \url{https://openreview.net/forum?id=u2Pd6x794I}
\BIBentrySTDinterwordspacing

\bibitem{JSD}
D.~Endres and J.~Schindelin, ``A new metric for probability distributions,''
  \emph{IEEE Transactions on Information Theory}, vol.~49, no.~7, pp.
  1858--1860, 2003.

\bibitem{ma2020normalized}
X.~Ma, H.~Huang, Y.~Wang, S.~Romano, S.~Erfani, and J.~Bailey, ``Normalized
  loss functions for deep learning with noisy labels,'' in \emph{ICML}, 2020.

\bibitem{bengio2009curriculum}
Y.~Bengio, J.~Louradour, R.~Collobert, and J.~Weston, ``Curriculum learning,''
  in \emph{Proceedings of the 26th annual international conference on machine
  learning}, 2009, pp. 41--48.

\bibitem{chen2020adversarial}
M.~Chen, S.~Zhao, H.~Liu, and D.~Cai, ``Adversarial-learned loss for domain
  adaptation,'' in \emph{Proceedings of the AAAI conference on artificial
  intelligence}, vol.~34, no.~04, 2020, pp. 3521--3528.

\bibitem{cui2020gradually}
S.~Cui, S.~Wang, J.~Zhuo, C.~Su, Q.~Huang, and Q.~Tian, ``Gradually vanishing
  bridge for adversarial domain adaptation,'' in \emph{Proceedings of the
  IEEE/CVF conference on computer vision and pattern recognition}, 2020, pp.
  12\,455--12\,464.

\bibitem{xia2021adaptive}
H.~Xia, H.~Zhao, and Z.~Ding, ``Adaptive adversarial network for source-free
  domain adaptation,'' in \emph{Proceedings of the IEEE/CVF international
  conference on computer vision}, 2021, pp. 9010--9019.

\bibitem{YM.2020Self-labelling}
\BIBentryALTinterwordspacing
A.~YM., R.~C., and V.~A., ``Self-labelling via simultaneous clustering and
  representation learning,'' in \emph{International Conference on Learning
  Representations}, 2020. [Online]. Available:
  \url{https://openreview.net/forum?id=Hyx-jyBFPr}
\BIBentrySTDinterwordspacing

\bibitem{zhang2021unsupervised}
H.~Zhang, Y.~Zhang, K.~Jia, and L.~Zhang, ``Unsupervised domain adaptation of
  black-box source models,'' \emph{arXiv preprint arXiv:2101.02839}, 2021.

\bibitem{liang2021source}
J.~Liang, D.~Hu, Y.~Wang, R.~He, and J.~Feng, ``Source data-absent unsupervised
  domain adaptation through hypothesis transfer and labeling transfer,''
  \emph{IEEE Transactions on Pattern Analysis and Machine Intelligence},
  vol.~44, no.~11, pp. 8602--8617, 2021.

\bibitem{saenko2010adapting}
K.~Saenko, B.~Kulis, M.~Fritz, and T.~Darrell, ``Adapting visual category
  models to new domains,'' in \emph{Computer Vision--ECCV 2010: 11th European
  Conference on Computer Vision, Heraklion, Crete, Greece, September 5-11,
  2010, Proceedings, Part IV 11}.\hskip 1em plus 0.5em minus 0.4em\relax
  Springer, 2010, pp. 213--226.

\bibitem{venkateswara2017deep}
H.~Venkateswara, J.~Eusebio, S.~Chakraborty, and S.~Panchanathan, ``Deep
  hashing network for unsupervised domain adaptation,'' in \emph{Proceedings of
  the IEEE conference on computer vision and pattern recognition}, 2017, pp.
  5018--5027.

\bibitem{peng2018visda}
X.~Peng, B.~Usman, N.~Kaushik, D.~Wang, J.~Hoffman, and K.~Saenko, ``Visda: A
  synthetic-to-real benchmark for visual domain adaptation,'' in
  \emph{Proceedings of the IEEE Conference on Computer Vision and Pattern
  Recognition Workshops}, 2018, pp. 2021--2026.

\bibitem{lin2014microsoft}
T.-Y. Lin, M.~Maire, S.~Belongie, J.~Hays, P.~Perona, D.~Ramanan,
  P.~Doll{\'a}r, and C.~L. Zitnick, ``Microsoft coco: Common objects in
  context,'' in \emph{Computer Vision--ECCV 2014: 13th European Conference,
  Zurich, Switzerland, September 6-12, 2014, Proceedings, Part V 13}.\hskip 1em
  plus 0.5em minus 0.4em\relax Springer, 2014, pp. 740--755.

\bibitem{he2016deep}
K.~He, X.~Zhang, S.~Ren, and J.~Sun, ``Deep residual learning for image
  recognition,'' in \emph{Proceedings of the IEEE conference on computer vision
  and pattern recognition}, 2016, pp. 770--778.

\bibitem{na2021fixbi}
J.~Na, H.~Jung, H.~J. Chang, and W.~Hwang, ``Fixbi: Bridging domain spaces for
  unsupervised domain adaptation,'' in \emph{Proceedings of the IEEE/CVF
  Conference on Computer Vision and Pattern Recognition}, 2021, pp. 1094--1103.

\bibitem{huang2021model}
J.~Huang, D.~Guan, A.~Xiao, and S.~Lu, ``Model adaptation: Historical
  contrastive learning for unsupervised domain adaptation without source
  data,'' \emph{Advances in Neural Information Processing Systems}, vol.~34,
  pp. 3635--3649, 2021.

\end{thebibliography}

\end{document}